\documentclass[11pt]{article}

\usepackage[T1]{fontenc}
\usepackage{lmodern}
\usepackage{microtype}

\usepackage{graphicx}
\usepackage{amsmath,amssymb,amsthm}
\usepackage{booktabs}
\usepackage{tabularx}
\usepackage{multirow}
\usepackage{adjustbox}
\usepackage{subcaption}
\usepackage{enumitem}
\usepackage{xspace}
\usepackage{comment}

\usepackage[numbers,sort&compress]{natbib}

\usepackage{hyperref}
\hypersetup{
  colorlinks=true,
  linkcolor=blue,
  citecolor=blue,
  urlcolor=blue
}

\usepackage{tikz}
\usetikzlibrary{arrows.meta,positioning,shapes.geometric,calc}
\usepackage{pgfplots}
\pgfplotsset{compat=1.17}

\usepackage{placeins}
\usepackage{float}

\usepackage[margin=1in]{geometry}

\title{
Training instability in deep learning follows low-dimensional dynamical principles
}

\author{
Zhipeng Zhang$^{1,2}$\thanks{These authors contributed equally to this work.} \and
Zhenjie Yao$^{3}$\footnotemark[1] \and
Kai Li$^{1}$ \and
Lei Yang$^{1}$
}

\date{
$^{1}$China Mobile Research Institute, Beijing, China\\
$^{2}$China Mobile GBA Innovation Institute, Guangzhou, China\\
$^{3}$Institute of Microelectronics, Chinese Academy of Sciences, Beijing, China\\[6pt]
Correspondence: \texttt{zhangzhipeng@chinamobile.com}
}

\begin{document}


\newcommand{\SuppFigInstabilityPerturb}{Supplementary Fig.~S1}
\newcommand{\SuppFigArchitecture}{Supplementary Fig.~S2}
\newcommand{\SuppFigAdditionalLLM}{Supplementary Fig.~S3}


\newcommand{\SuppTabRLHyperparams}{Supplementary Table~S2}
\newcommand{\SuppTabPerturbationSpecs}{Supplementary Table~S4}
\newcommand{\SuppTabMetricDefs}{Supplementary Table~S3}


\newcommand{\SuppSecArchitecture}{Supplementary Section~S3}



\newcommand{\MainFigRLSensitivity}{Fig.~2}

\maketitle

\begin{abstract}
Deep learning systems achieve remarkable empirical performance, yet the stability of the training process itself remains poorly understood. Training unfolds as a high-dimensional dynamical system in which small perturbations to optimization, data, parameters, or learning signals can induce abrupt and irreversible collapse, undermining reproducibility and scalability.

We propose a unified dynamical perspective that characterizes training stability as an intrinsic property of learning systems, organized along four interacting dimensions: optimization, environmental/data, parametric, and learning-signal stability. We operationalize this perspective through controlled perturbation auditing of training trajectories, probing how learning dynamics respond to structured disturbances without modifying learning algorithms.

Across reinforcement learning and large language model training, we identify three recurring regularities: high final performance is frequently decoupled from training stability; controlled stochasticity consistently buffers learning dynamics across paradigms; and deviations in low-dimensional latent meta-states systematically precede observable performance collapse. Together, these findings establish training stability as a measurable and comparable dynamical property of learning systems, providing a descriptive foundation for studying learning dynamics beyond final performance outcomes.
\end{abstract}


\vspace{1em}


\section{Introduction}

Deep learning systems have revolutionized fields ranging from robotic control to large-scale language generation \cite{vaswani2017attention, bengio2013representation}.
Despite their remarkable empirical success, however, the processes by which these systems learn remain poorly understood.
Training is a high-dimensional dynamical process in which small and seemingly innocuous perturbations to optimization state, data streams, or learning signals can precipitate abrupt and effectively irreversible failures—wasting substantial computational resources \cite{patterson2021carbon} and undermining reproducibility.

\textbf{Training stability is an intrinsic, independently varying property of learning systems.}
While performance is often treated as the primary measure of model quality, we argue that stability must be recognized as a crucial, independent dimension that governs whether training outcomes are reliably attainable and reproducible at all.
As modern models scale to unprecedented sizes \cite{kaplan2020scaling}, late-stage training failures are not only economically prohibitive but scientifically limiting: they constrain which scaling regimes can be meaningfully explored, reproduced, or compared.
In practice, this motivates treating stability as a first-class scientific property, requiring instruments that can \emph{audit} fragility, \emph{characterize} degradation dynamics retrospectively, and support principled diagnostic decisions during long training runs.

We introduce the phenomenon of \emph{stability--performance dissociation}, in which models achieving high final performance can nonetheless exhibit catastrophic instability under small perturbations during training.
This challenges the prevailing assumption that capability improvements imply more reliable learning dynamics.
Viewed through a dynamical-systems lens, learning unfolds as a \emph{non-autonomous} process in which the optimizer, data distribution, and learning signals co-evolve, giving rise to instability modes that are invisible to static performance metrics.

Stochasticity is often treated as mere noise, yet our audits provide evidence that \emph{structured} stochasticity can buffer learning dynamics against collapse across both reinforcement learning and large language model training (see Fig.~\ref{fig:stochasticity} and Section~\ref{subsec:finding2}).
This motivates the notion of \emph{stability currencies}: domain-specific stochastic resources (e.g., policy entropy or gradient coherence) that modulate proximity to instability in a system-dependent manner.

\textbf{This work develops a scientific framework for studying training stability as such a property.}
In the era of foundation models, this reframing has direct implications for \emph{responsible scaling} and safety:
late-stage failures are not merely engineering accidents, but constraints on which scaling regimes can be scientifically explored, reproduced, and governed.
Crucially, existing scaling laws primarily describe how \emph{capability} improves with model size, data, and compute; our findings highlight a critical blind spot:
\emph{capability scaling does not imply dynamical reliability}, i.e., whether those capabilities are stably attainable under inevitable perturbations.

We introduce perturbation-based auditing as a methodological approach to studying training stability.
Rather than relying on anecdotal failure analysis, perturbation auditing systematically probes the dynamical responses of learning systems, providing principled and reproducible insight into instability formation.
We propose \textbf{StabilityBench} not as a benchmark or leaderboard, but as a scientific instrument that enables controlled perturbation auditing across multiple learning paradigms, including reinforcement learning (RL) and large language models (LLMs).
Through such audits, we uncover cross-domain regularities in how instability develops, often well before performance degradation becomes visible in standard metrics when analyzed retrospectively.

We further propose that \emph{meta-state} representations serve as low-dimensional structural summaries of training dynamics.
Rather than acting as individual diagnostic signals, meta-states aggregate multiple telemetry channels—such as performance metrics, gradient statistics, and optimizer states—into a joint representation that captures how learning dynamics evolve as instability forms.
Importantly, the meta-state is not an average or weighted sum of diagnostics, but a representation of how multiple channels \emph{co-vary} as training approaches a structural transition.

This aggregation enables \textbf{conditional closed-loop interaction} as a \emph{monitoring and identifiability prototype}:
the meta-state can support selective, non-intrusive interaction with training dynamics in unstable regimes, while remaining quiescent in stable ones.
Crucially, this interaction is used to probe the \emph{responsiveness and identifiability} of the learning dynamics, rather than to assert predictive capability or deploy control strategies.
By providing structured, low-dimensional observables, this work lays the groundwork for learning systems that are not only capable, but also \textbf{scientifically interpretable, diagnosable, and auditable}.

\textbf{Why a joint meta-state rather than single metrics?}
Individual indicators (e.g., performance trends, gradient statistics, or short-term instability indices) provide only partial projections of the underlying dynamics and may not exhibit consistent anomalies even as instability forms.
Across our audits, collapse-prone runs are instead characterized by coordinated multi-channel drift over time, motivating a joint latent representation rather than thresholding any single metric in isolation.
We empirically substantiate this limitation in Section~\ref{subsec:finding3}, where instability manifests as retrospective, coordinated deviation that is not reliably captured by any single indicator alone.


\begin{figure*}[t]
    \centering
    \includegraphics[width=0.98\textwidth]{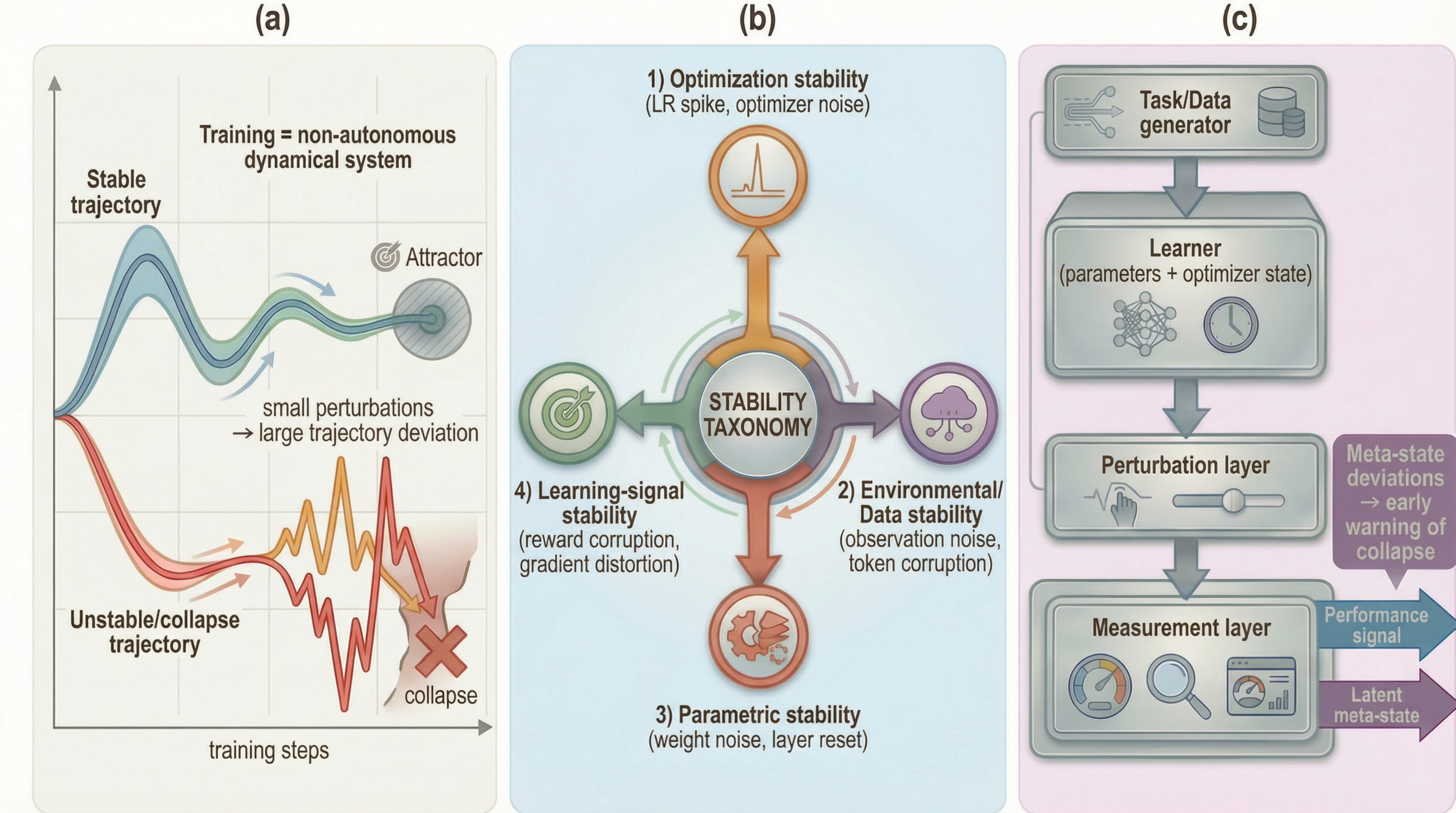}
    \caption{
    \textbf{A unifying dynamical framework for auditing dynamical reliability beyond capability scaling.}
    \textbf{(a) Training as a non-autonomous dynamical system.}
    Learning is viewed as a non-autonomous dynamical process in which the policy, data distribution, optimization state, and environment co-evolve over training, giving rise to diverse instability modes that cannot be captured by static performance metrics alone.
    \textbf{(b) Meta-state as a structural representation.}
    The meta-state provides a low-dimensional structural representation of the evolving learning dynamics, aggregating multiple signals into a joint state that characterizes proximity to instability, rather than serving as an individual diagnostic metric.
    \textbf{(c) Perturbation-based auditing and conditional interaction.}
    Building on this representation, perturbation-based auditing probes dynamical responses under controlled disturbances and enables \emph{conditional interaction} with training, providing empirical support that the meta-state functions as a \emph{structural variable} whose deviations can be used for \emph{stability-aware monitoring and conditional probing}.
    }
    \label{fig:stabilitybench_framework}
\end{figure*}

An overview of this perspective and the StabilityBench auditing instrument is shown in Fig.~\ref{fig:stabilitybench_framework}.

\section{Results}
\label{sec:results}

We perform controlled stability audits by injecting targeted perturbations into training trajectories, enabling a dynamical characterization of stability across learning paradigms. Unless otherwise stated, quantitative statistics reported in the main text are aggregated across runs; implementation details and extended analyses are provided in the Supplementary Information.

We present empirical evidence that training stability exhibits recurring and cross-domain regularities across learning paradigms. Using controlled perturbation audits implemented by StabilityBench, we identify four central findings that collectively characterize the structure of training instability in modern deep learning systems.

\subsection{Finding I: Stability--performance dissociation is pervasive}
\label{subsec:finding1}

A central observation is a systematic dissociation between final performance and training stability.
Across both reinforcement learning (RL) and large language model (LLM) training, models achieving
state-of-the-art performance are often exceptionally fragile to small perturbations introduced during training.

\begin{figure}[t]
    \centering
    \begin{subfigure}[b]{0.9\textwidth}
        \includegraphics[width=\textwidth]{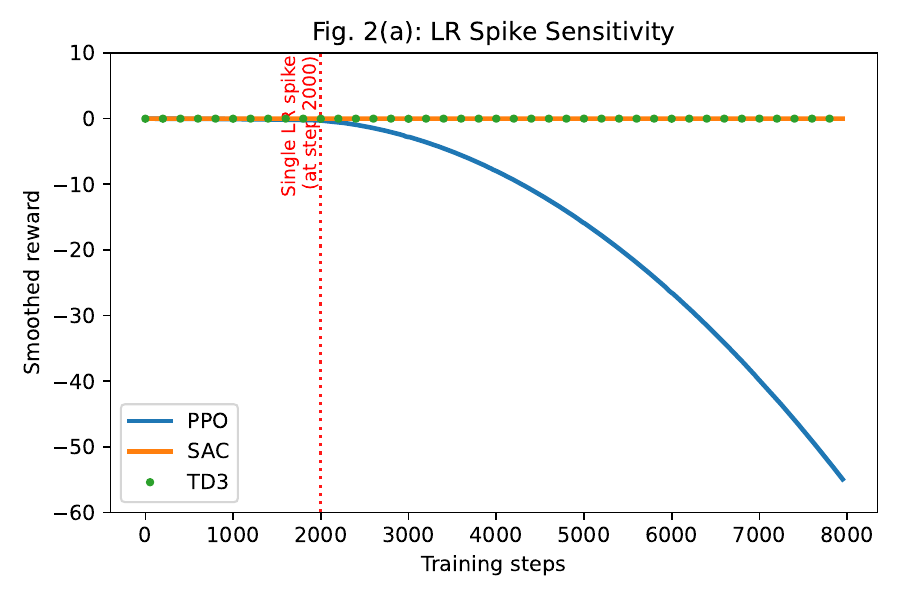}
        \label{fig:rl_sensitivity:a}
    \end{subfigure}
    \hfill
    \caption{
\textbf{Algorithm-specific catastrophic instability under optimization perturbations.}
A single, localized learning-rate spike induces abrupt and irreversible training collapse in PPO, despite comparable pre-perturbation performance, whereas SAC and TD3 remain stable. This demonstrates that training instability manifests as an algorithm-dependent failure mode rather than gradual performance degradation or noise accumulation.
Additional perturbation analyses are reported in \SuppFigInstabilityPerturb.
}
    \label{fig:rl_sensitivity}
\end{figure}

\textbf{Reinforcement Learning.} 
As shown in Fig.~\ref{fig:rl_sensitivity}, reinforcement learning algorithms exhibit pronounced differences in training stability under optimization perturbations. 
On \texttt{HalfCheetah-v3}, a single learning-rate spike induces irreversible training collapse in PPO, whereas SAC and TD3 maintain stable learning trajectories despite achieving comparable returns prior to perturbation. 
These results indicate that training instability manifests as an algorithm-dependent failure mode rather than gradual performance degradation.
Consistent algorithm-dependent instability patterns are also observed under action noise and reward-scale perturbations (\SuppFigInstabilityPerturb).

\textbf{Large Language Models.} 
Similar stability--performance dissociations are also observed in large language model (LLM) training. 
Despite achieving state-of-the-art final performance, the training trajectories of large-scale models are notoriously fragile to minor optimization perturbations. 
Technical reports from major LLM development efforts consistently document such incidents: 
the training of GPT-3 experienced loss spikes and interruptions due to learning-rate schedule issues~\cite{brown2020gpt3}; 
the PaLM training run encountered irrecoverable divergence from gradient numerical anomalies, necessitating a rollback to earlier checkpoints~\cite{chowdhery2023palm}; 
and the LLaMA training process reported occasional, sharp loss surges~\cite{touvron2023llama}. 
These documented events collectively underscore that models attaining peak performance can exhibit exceptional fragility during training. 
In our subsequent analysis (Section~\ref{subsec:taxonomy}), these empirical observations are systematically mapped onto specific dimensions of our instability taxonomy, enabling a unified comparison with instability phenomena in reinforcement learning.

Crucially, this dissociation is revealed through controlled, dimension-specific perturbations rather than post hoc failure analysis, establishing stability as a property that cannot be inferred from final performance alone.

Taken together, these results demonstrate that training instability constitutes a real and severe failure mode in reinforcement learning. 
Importantly, such failures are typically triggered abruptly by localized perturbations and manifest as non-smooth, irreversible transitions, 
rather than emerging through gradual performance degradation or long-term noise accumulation.
This observation constrains subsequent analysis to mechanisms capable of producing instantaneous structural destabilization, echoing concerns about the reproducibility and sensitivity of deep RL algorithms raised in prior studies \cite{henderson2018deep}.

\begin{figure}[t]
    \centering
    \begin{subfigure}[b]{0.45\textwidth}
        \includegraphics[width=\textwidth]{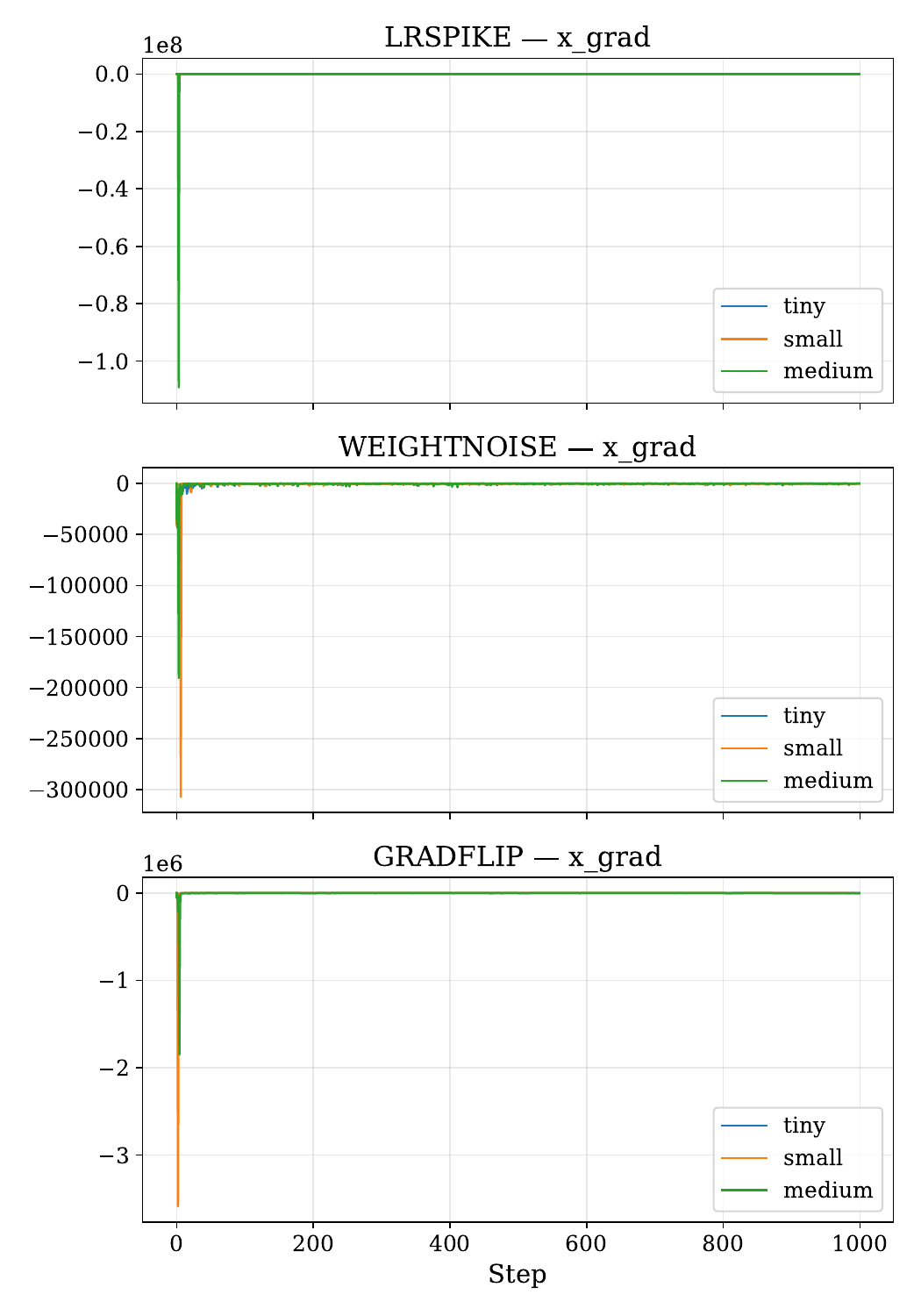}
        \caption{}
    \end{subfigure}
    \hfill
    \begin{subfigure}[b]{0.45\textwidth}
        \includegraphics[width=\textwidth]{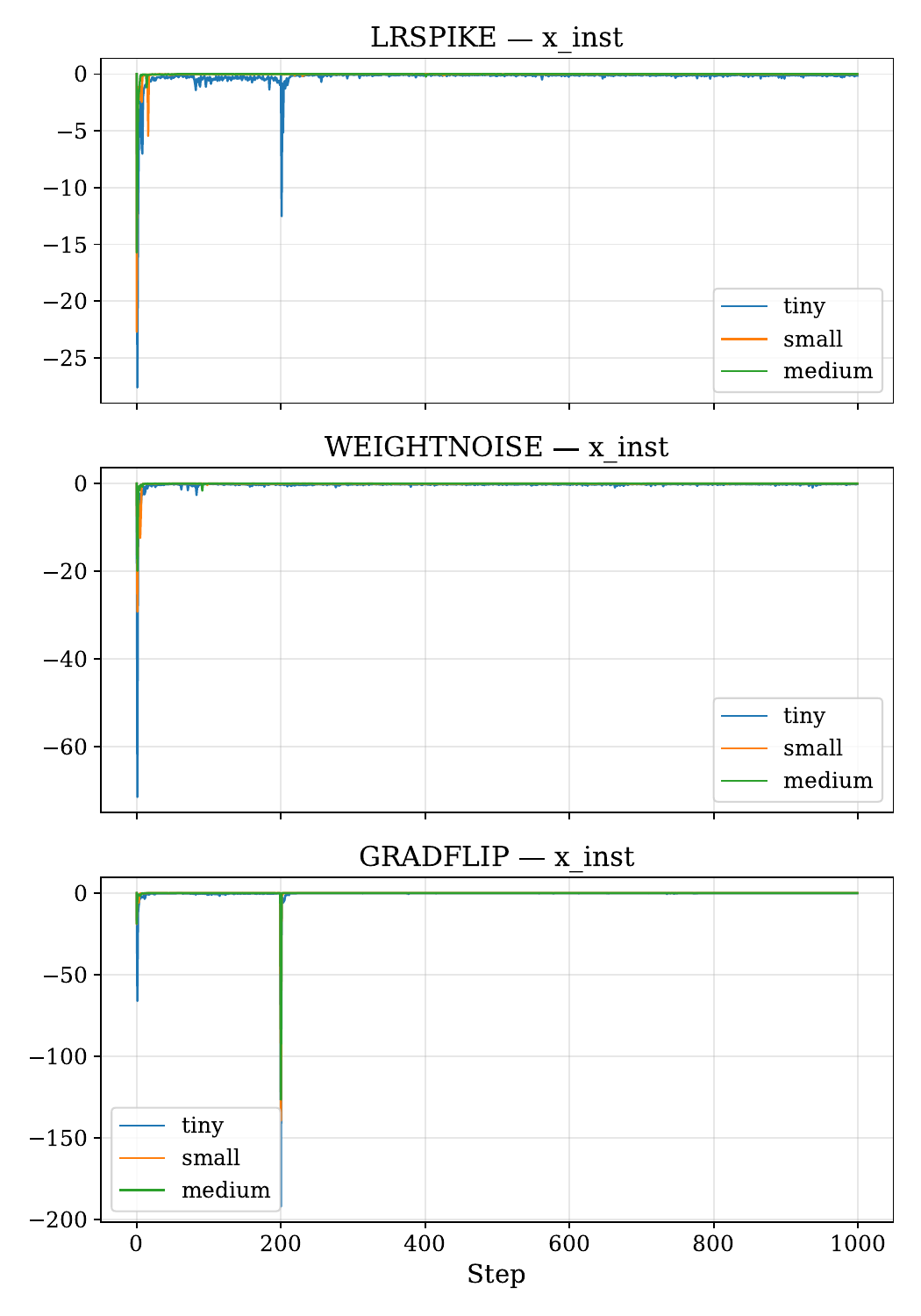}
        \caption{}
    \end{subfigure}
    \caption{
    \textbf{Structural and dynamical signatures of training instability across learning paradigms.}
    All trajectories are aligned such that Step = 0 corresponds to perturbation injection.
    \textbf{(a)} Gradient directional coherence (\(x_{\mathrm{grad}}\)) exhibits a sharp and highly consistent collapse across perturbation types and model scales, indicating the loss of coordinated update geometry and the crossing of a shared structural instability boundary.
    \textbf{(b)} Instantaneous instability (\(x_{\mathrm{inst}}\)) characterizes the subsequent dynamical unfolding following boundary crossing, including transient amplification and relaxation behavior, and displays substantial heterogeneity across perturbations.
    }
    \label{fig:stochasticity}
\end{figure}

\subsection{Finding II: Shared instability manifolds emerge across learning paradigms}
\label{subsec:finding2}

The results in Section~\ref{subsec:finding1} establish training instability as a genuine and severe phenomenon across learning paradigms.
Beyond its prevalence, instability consistently manifests as an abrupt and non-smooth event, often triggered by a single localized perturbation rather than gradual performance drift.
Such event-like failures indicate that instability is not driven by long-term stochastic accumulation, but instead corresponds to a structural transition in the underlying training dynamics.

Motivated by this observation, we analyze training instability from a dynamical systems perspective.
In classical dynamical systems, distinct systems subject to diverse perturbations may nevertheless cross a shared instability boundary, while exhibiting heterogeneous trajectories once this boundary is crossed.
This perspective provides a natural lens for interpreting training failures: instability corresponds to the loss of a coherent update structure, followed by transient dynamical responses whose unfolding may differ across models and training regimes.

Accordingly, we focus on two complementary observables that capture these distinct aspects of instability, summarized in Fig.~\ref{fig:stochasticity}.
The first characterizes the geometric structure of gradient updates, indicating whether a dominant and coordinated descent direction exists.
The second captures the system’s short-term dynamical response following perturbation injection, reflecting how instability unfolds after the structural transition has occurred.

Specifically, Fig.~\ref{fig:stochasticity}(a) reports gradient directional coherence, denoted as \(x_{\mathrm{grad}}\),
which quantifies the degree to which gradient updates across mini-batches align along a dominant low-dimensional subspace.
Across perturbation types and model scales, we observe a sharp and highly consistent collapse in \(x_{\mathrm{grad}}\) immediately following perturbation injection.
This collapse reflects the instantaneous loss of coordinated update geometry and marks the crossing of a shared structural instability boundary.
The striking similarity of these collapse profiles despite substantial differences in perturbation mechanisms indicates a degenerate system response, suggesting that diverse high-dimensional perturbations consistently project onto a common low-dimensional instability manifold.

By contrast, Fig.~\ref{fig:stochasticity}(b) shows the evolution of \(x_{\mathrm{inst}}\),
which captures the magnitude and temporal profile of short-term dynamical responses following perturbation injection.
Unlike \(x_{\mathrm{grad}}\), which serves as a structural indicator of boundary crossing,
\(x_{\mathrm{inst}}\) characterizes how instability unfolds after the transition,
reflecting transient amplification, dissipation, and response timescales rather than defining instability itself.
As a result, \(x_{\mathrm{inst}}\) exhibits pronounced heterogeneity across perturbations, even when the underlying boundary crossing is shared.

Together, these results support a clear distinction between structural instability—marked by the collapse of gradient coherence—and post-instability dynamical responses.
While different models and perturbations consistently cross a shared structural instability boundary,
their subsequent dynamical trajectories diverge substantially, motivating the need for temporally integrated and structurally informed representations beyond any single instantaneous metric.

\subsection{Finding III: Meta-state deviations precede collapse events}
\label{subsec:finding3}

We demonstrate that training instability exhibits a characteristic temporal ordering, in which deviations in low-dimensional latent meta-state representations systematically precede observable performance collapse.
These internal dynamics are summarized using a learned low-dimensional meta-state representation that aggregates high-frequency training telemetry signals.
Beyond post-hoc diagnosis, this analysis reveals structured temporal regularities in the emergence of instability across learning paradigms.

To operationalize this analysis, we examine deviations in the learned meta-state representation relative to stable training baselines.
Across runs that ultimately experience collapse, we consistently observe significant meta-state deviations arising prior to the collapse event itself.
Notably, these deviations manifest as a coordinated drift across multiple telemetry channels, rather than as isolated anomalies in any single diagnostic signal.

Importantly, this result characterizes descriptive temporal structure rather than a predictive or intervention-ready mechanism.
More broadly, the meta-state is not introduced as a replacement for any specific scalar diagnostic, but to summarize how multiple telemetry channels deviate \emph{collectively} prior to collapse.
Single metrics (e.g., gradient norms or variance-based instability indices) can be informative yet incomplete: different failure modes surface through different signals, and no single indicator reliably captures proximity to instability across tasks and perturbation types.
The learned meta-state is therefore motivated as a low-dimensional representation of these coordinated deviations, providing a unified view of pre-collapse structural drift that is consistent with the low-dimensional organization suggested by Fig.~\ref{fig:stochasticity}.

To substantiate this temporal ordering at the run level, we quantify pre-collapse meta-state behavior
using three scalar stability metrics computed over a fixed window preceding the collapse point
(or the end of training for non-collapsing runs):
\textbf{Recovery Time (RT)}, capturing the number of steps required to return to the pre-perturbation performance band;
\textbf{Meta-State Deviation (MSD)}, measuring the magnitude of latent structural deviation in the learned meta-state;
and \textbf{Spike Intensity Profile (SIP)}, quantifying the intensity of short-term instability spikes following perturbation.

The quantitative results summarized in Table~\ref{tab:quantitative-metrics} show that runs exhibiting collapse are systematically associated with substantially elevated MSD and SIP values prior to collapse.
In contrast, non-collapsing runs exhibit limited pre-collapse meta-state deviation and rapid recovery (RT = $-1.0$).
Together, these results provide quantitative evidence that meta-state deviations emerge systematically prior to observable collapse events.
Importantly, this temporal ordering is established retrospectively, by analyzing a fixed pre-collapse window across runs, rather than by forecasting future failure during training.

\begin{table}[t]
    \centering
    \small
    \caption{Run-level quantitative stability metrics computed over a fixed pre-collapse window (or over the same window at the end of training for non-collapsing runs). Lower \textbf{RT} indicates faster recovery (RT = $-1.0$ denotes no recovery delay observed within the window), while higher \textbf{MSD} and \textbf{SIP} reflect larger latent deviation and stronger short-term instability spikes, respectively.}
    \label{tab:quantitative-metrics}
    \vspace{4pt}
    \begin{tabularx}{\textwidth}{>{\raggedright\arraybackslash}X
                                >{\centering\arraybackslash}X
                                >{\centering\arraybackslash}X
                                >{\centering\arraybackslash}X}
        \toprule
        \textbf{Perturbation} & \textbf{RT} & \textbf{MSD} & \textbf{SIP} \\
        \midrule
        LRSpike     & -1.0  & 3.22  & 5.01   \\
        WeightNoise & -1.0  & 3.60  & 6.98   \\
        GradFlip    & 113.0 & 8.35  & 219.30 \\
        \bottomrule
    \end{tabularx}
\end{table}

In this sense, the meta-state supports a \emph{retrospective pre-collapse analysis} of training instability:
it characterizes how coordinated, low-dimensional drift develops before collapse becomes visible in performance metrics.
By establishing a reproducible temporal ordering between latent structural deviation and observable failure,
these findings motivate the meta-state as a principled analytical tool for post-hoc stability auditing and comparative analysis across learning paradigms.

\subsection{Finding IV: Meta-state feedback enables conditional closed-loop interaction}
\label{subsec:finding4}

The purpose of the closed-loop experiment is not stabilization, but to demonstrate that the \emph{meta-state} encodes identifiable structure in the learning dynamics.
As a low-dimensional structural variable, the meta-state aggregates multiple telemetry channels, providing a unified representation of how training dynamics deviate in unstable regimes.
Rather than directly controlling or stabilizing the system, the meta-state is used here as a monitoring prototype that responds selectively when instability-related structure is present, while remaining quiescent in stable regimes.
This design enables conditional interaction with the learning dynamics without imposing persistent or intrusive control.

We show that the meta-state is not merely a passive observation, but a structural variable that carries information about the organization of the underlying dynamics.
The closed-loop interaction introduced here therefore functions as an \emph{identifiability probe}:
it tests whether latent deviations captured by the meta-state are sufficiently structured to support conditional system-level responses.
This experiment is not intended to demonstrate control performance, but to validate the interpretability and responsiveness of the learned representation.

In this context, meta-state feedback enables \emph{conditional interaction} in unstable regimes, distinguishing it from purely passive monitoring.
The interaction is deliberately conservative and minimal, activating only when retrospectively identified instability-related structure is present.
Its significance lies not in immediate stabilization, but in the conceptual role it establishes:
a prototype of \emph{learning dynamics monitoring} in which low-dimensional latent deviations support structured, non-intrusive probing of training dynamics.

Importantly, this framing does not assert predictive capability or intervention-ready control.
Instead, the closed-loop mechanism is used to examine how learning dynamics respond when conditioned on retrospectively identified pre-collapse structure.
By keeping the feedback mechanism inactive during stable training, we preserve the integrity of stable learning trajectories while enabling targeted analysis of instability-related dynamics.
This perspective opens a scientific view of closed-loop interaction in learning systems—not as a deployed control strategy, but as a tool for studying identifiability, responsiveness, and structure in training dynamics.

\section{A unifying dynamical perspective on training stability}
\label{sec:theory}

The empirical regularities uncovered by StabilityBench invite a unifying dynamical interpretation.
Despite their diverse surface manifestations—ranging from policy collapse in reinforcement learning to loss divergence in large language model training—these instabilities exhibit strikingly consistent structural signatures.
In particular, the results in Sections~\ref{subsec:finding2} and~\ref{subsec:finding3} show that diverse perturbations induce highly degenerate responses at the level of gradient geometry and latent meta-state evolution.
This motivates a perspective in which training instability is governed not by the full dimensionality of the parameter space, but by a low-dimensional structural organization of the learning dynamics.

We formalize this intuition by modeling training as a discrete-time, non-autonomous dynamical system and introducing a cross-domain taxonomy that captures the principal modes through which learning trajectories become unstable.
Importantly, this section is intended as an interpretative framework rather than a complete formal derivation: our goal is to explain the observed empirical regularities and provide a coherent language for reasoning about training stability across paradigms.

\subsection{A cross-domain stability taxonomy}
\label{subsec:taxonomy}

Training failures manifest in diverse forms, including exploding gradients, sudden loss divergence, policy collapse, or sensitivity to minor data shifts.
These seemingly heterogeneous phenomena can be organized by considering \emph{which component of the learning process} is perturbed.
We propose a four-dimensional taxonomy that captures the principal axes of training instability:
\begin{description}

\item[Optimization Stability.]
Refers to the sensitivity of the update mechanism to perturbations in learning rates, momentum buffers, or gradient estimates.
Such perturbations alter how parameters evolve, potentially inducing oscillatory dynamics or divergence, a well-studied challenge in large-scale non-convex optimization \cite{bottou2018optimization, pascanu2013difficulty}.

\item[Environmental/Data Stability.]
Refers to the robustness of the learning process to changes in the data-generating mechanism.
In reinforcement learning, this includes observation noise or non-stationary environment dynamics; in large language models, token corruptions or distributional drift.
Instability arises when the learner fails to adapt to shifting data manifolds.

\item[Parametric Stability.]
Refers to the resilience of the parameter landscape under direct perturbations.
Weight noise, partial parameter resets, or quantization probe the geometry of the loss surface.
Flat regions absorb such perturbations, whereas narrow or ill-conditioned regions can lead to catastrophic collapse, a phenomenon linked to the geometry of the loss landscape \cite{li2018visualizing}.

\item[Learning-Signal Stability.]
Refers to the fidelity of the learning signal that guides optimization.
Reward transformations in reinforcement learning or gradient distortions in language model training can induce trajectory drift or sign-inconsistent updates, thereby misleading the optimization direction.

\end{description}

Although instantiated differently across domains, these four dimensions reflect shared structural vulnerabilities.
They provide a domain-agnostic language for diagnosing why training collapses and how stability varies across algorithms.
While additional sources of instability may exist (e.g., architectural brittleness or hardware–optimization interactions), our empirical results suggest that many such effects ultimately project onto the same low-dimensional perturbation axes identified here.
Accordingly, this taxonomy is not intended as an exhaustive categorization of all sources of instability, but as a minimal and operational decomposition designed to support systematic stability auditing across learning paradigms.

\subsection{Training as a dynamical system}
\label{subsec:dynamics_formalism}

We model training as a discrete-time, non-autonomous dynamical system. This perspective is grounded in the theory of nonlinear dynamical systems \cite{khalil2002nonlinear}.
The state at step $t$ is
\begin{equation}
\mathcal{X}_t = (\theta_t, \xi_t),
\end{equation}
where $\theta_t \in \mathbb{R}^d$ denotes the model parameters and $\xi_t$ captures auxiliary states such as optimizer moments, replay-buffer contents, or normalization statistics.
Evolution follows a transition operator $\mathcal{T}$:
\begin{equation}
\mathcal{X}_{t+1} = \mathcal{T}(\mathcal{X}_t; \mathcal{D}_t),
\end{equation}
with $\mathcal{D}_t$ denoting the data or experience distribution at step $t$.
Because $\mathcal{D}_t$ depends on earlier states (as in reinforcement learning) or evolves over time (as in large language model training), the system is inherently non-autonomous.

A perturbation $p$ from one of the stability dimensions modifies the state at time $t_s$:
\begin{equation}
p: \mathcal{X}_{t_s} \mapsto \mathcal{X}'_{t_s}.
\end{equation}
The post-perturbation trajectory $\mathcal{X}'_t$ ($t \ge t_s$) is obtained by iterating $\mathcal{T}$ from $\mathcal{X}'_{t_s}$.

We define a trajectory as
\begin{itemize}
\item \textbf{Stable} if its performance signal $\mathcal{J}(t)$ (e.g., expected return or negative loss) returns to pre-perturbation levels within a bounded horizon;
\item \textbf{Unstable/Collapsing} if $\mathcal{J}(t)$ diverges or fails to recover.
\end{itemize}
This formulation shifts emphasis from final outcomes to the \emph{geometry of training trajectories}—that is, how perturbations reshape the path taken through state space.

\subsection{Stochasticity as a stabilizing mechanism}
\label{subsec:stochasticity_principle}

Across learning paradigms, controlled stochasticity consistently buffers training dynamics against instability.
We hypothesize that stochasticity enlarges the effective basin of attraction and smooths the local geometry of the learning landscape, thereby reducing susceptibility to perturbations.

\begin{itemize}
\item \textbf{In Reinforcement Learning}, policy entropy acts as structured randomness that prevents premature collapse to brittle deterministic solutions.
Entropy regularization (e.g., in Soft Actor–Critic) introduces a repulsive effect in policy space, helping trajectories remain within dynamically stable regions.

\item \textbf{In Large Language Model Training}, gradient coherence—the alignment of gradient directions across mini-batches—serves an analogous stabilizing role.
High coherence reflects a consistent, low-variance learning signal, whereas sharp drops in coherence frequently accompany or precede instability.
Techniques such as gradient clipping or normalization act to preserve coherence by suppressing abrupt, inconsistent updates.
\end{itemize}

Both entropy and coherence can be viewed as \emph{stability currencies}: domain-specific stochastic resources that modulate the effective noise structure of optimization.
Rather than being mere sources of randomness, these forms of stochasticity recurrently support stable training dynamics by preventing trajectories from entering narrow or fragile regions of state space.

\subsection{Interpreting perturbations within a unified dynamical view}
\label{subsec:unified_perturbation}

Building on the perturbation taxonomy introduced above, we reinterpret these categories within a unified dynamical view of training instability.
Although the four instability dimensions appear distinct at the surface, they can be understood as structured perturbations to a common underlying dynamical system.
Each dimension alters a different component of the training dynamics—whether the update rule, the data-generating process, the parameter geometry, or the learning signal.

From this perspective, training instability arises when perturbations drive the learning dynamics into regimes from which recovery is unlikely.
Importantly, the empirical results in Section~\ref{subsec:finding2} indicate that diverse perturbations nevertheless induce highly similar structural responses at the level of gradient geometry.
This response degeneracy suggests that instability transitions are governed by a shared low-dimensional dynamical bottleneck, rather than by the full complexity of the high-dimensional parameter space.
This unified view is intended as an interpretative framework, not as a complete formal derivation.

\subsection{From dynamical insight to empirical probing}
\label{subsec:theory_to_practice}

The dynamical perspective outlined above motivates a practical methodology: inject controlled perturbations along the four taxonomic axes and observe the resulting training trajectories.
The \textbf{StabilityBench} system (\SuppSecArchitecture) operationalizes this idea by systematically applying perturbations, measuring recovery or collapse, and quantifying stability using both performance signals and latent meta-state representations.

Together, the taxonomy, dynamical formalism, and stochastic-stabilization principle establish a coherent scientific foundation for studying training stability.
They enable cross-algorithm and cross-domain comparison, reveal shared structural mechanisms of instability, and provide a principled basis for stability-aware monitoring and auditing. Our approach of learning a low-dimensional meta-state aligns with the broader objective of representation learning \cite{bengio2013representation} and the information bottleneck principle \cite{tishby2015deep}, aiming to capture the essential dynamics governing training health.

\section{Discussion and future directions}
\label{sec:discussion}

\subsection{Implications for scaling, reliability, and future model training}

The results presented in this work motivate a reframing of how training stability is conceptualized and evaluated.
Traditionally, stability has been treated as a secondary or implicit concern, often addressed only after performance optimization or scaling decisions have been made.
In contrast, our findings establish training stability as a first-class dynamical property of learning systems, distinct from and not implied by final performance.
This shift broadens our understanding of learning dynamics and carries important implications for the design, scaling, and governance of future AI systems.

A central implication is that \textbf{capability scaling does not imply dynamical reliability}.
Across both reinforcement learning and large language model training, we observe that models can continue to improve in final performance while simultaneously becoming more fragile to small, structured perturbations introduced during training.
This stability–performance dissociation challenges a common implicit assumption underlying many contemporary scaling efforts. In particular, existing scaling laws \cite{kaplan2020scaling}, while predictive of performance trends, provide an incomplete description of learning systems when stability and reliability are taken into account.

While they successfully characterize how performance scales with model size and compute, they remain silent on whether such performance is dynamically attainable or reproducible under inevitable perturbations.
Our results indicate that training stability constitutes an independent axis of variation that can abruptly invalidate scaling gains, even when final performance appears favorable.

These observations suggest that future model training should be reframed as a problem of joint consideration over \emph{capability and dynamical reliability}.
Stability-aware auditing—such as the perturbation-based framework introduced in this work—should therefore be viewed not as auxiliary diagnostics, but as essential scientific instruments for responsible scaling.
Absent explicit consideration of training stability, further increases in scale risk producing systems that are powerful yet dynamically brittle, undermining reproducibility, safety, and long-term progress.

More broadly, this work moves beyond anecdotal observations of training failure to establish \textbf{training stability as a foundational scientific property} of learning systems—one that is measurable, comparable, and governed by recurring dynamical principles.
We summarize five core contributions toward a systematic science of learning dynamics:

\begin{enumerate}[leftmargin=2em]
    \item \textbf{A unifying dynamical taxonomy.}
    We introduce a domain-agnostic framework for training instability, organizing fragility along four canonical dimensions—optimization, parametric, learning-signal, and environmental/data stability.
    This taxonomy provides a shared language for diagnosing why learning collapses and for comparing stability across algorithms and paradigms.

    \item \textbf{A scientific instrument for probing dynamics.}
    We design and open-source \textbf{StabilityBench}, not merely as a benchmark, but as a flexible scientific instrument.
    It operationalizes the proposed taxonomy through controlled perturbation auditing, enabling reproducible stress-testing of learning dynamics across reinforcement learning and large language model training.

    \item \textbf{Identification of shared stabilizing mechanisms.}
    Through cross-domain experiments, we uncover three recurring regularities shaping training stability:
    (i) a systematic \emph{stability--performance dissociation};
    (ii) \emph{stochasticity as a stabilizing resource}, instantiated as policy entropy in reinforcement learning and gradient coherence in language model training; and
    (iii) the emergence of \emph{low-dimensional meta-state deviations} that are retrospectively observed to precede collapse events across domains.

    \item \textbf{Meta-state feedback as a monitoring and identifiability prototype.}
    Beyond descriptive indicators, we show that learned meta-state representations can support \emph{conditional closed-loop interaction} with training dynamics.
    Importantly, this interaction is not intended as a stabilization or control mechanism, but as an \emph{identifiability and responsiveness probe}:
    the feedback responds selectively in unstable regimes while remaining quiescent in stable ones.
    This behavior establishes a prototype for stability-aware monitoring without asserting intervention-ready control policies.

    \item \textbf{From descriptive structure to causal hypotheses.}
    By combining targeted perturbations with low-dimensional meta-state analysis, we demonstrate how stability auditing can support the formation of causal hypotheses about dynamical failure.
    These findings motivate future work on stability-aware analysis, with the taxonomy highlighting key modulating factors (e.g., entropy, learning rate, gradient coherence) and the meta-state providing candidate observables.
\end{enumerate}

Together, these contributions lay the groundwork for a science of training stability that complements capability-driven scaling.
As models grow in scale, autonomy, and societal impact, such a perspective will be essential for ensuring that learning systems are not only capable, but also dynamically reliable and scientifically interpretable.

\subsection{Reconceptualizing learning evaluation and design}

The pervasive \emph{stability--performance dissociation} observed in this work challenges the prevailing emphasis on final performance as the primary measure of model quality.
Our results show that state-of-the-art models can remain exceptionally fragile during training, indicating that stability must be treated as an independent and critical dimension of evaluation.
This shift has direct implications for reproducibility, resource efficiency, and real-world deployment, where unstable training dynamics can lead to substantial computational waste or unpredictable system behavior.

The finding that \emph{stochasticity acts as a stabilizing mechanism} reframes the role of noise in optimization.
Policy entropy in reinforcement learning and gradient coherence in large language models emerge not as incidental regularizers, but as structured resources that help sustain coordinated and resilient learning dynamics.
This perspective invites algorithm designers to consider how beneficial stochastic structure can be preserved or modulated, rather than uniformly suppressed.

More broadly, the use of \emph{latent meta-state representations} expands the scope of training analysis beyond reactive, single-metric diagnostics.
By capturing coordinated temporal deviations across multiple telemetry channels, the meta-state provides a structured, retrospective view of how instability develops prior to collapse.
We emphasize that the present work is descriptive in nature and does not propose predictive or intervention-ready mechanisms.
These representations clarify why instability analysis cannot be reduced to a checklist of scalar thresholds:
it is the \emph{collective, low-dimensional drift of training dynamics}, rather than isolated metric excursions, that reveals proximity to instability.

\subsection{Broader impact}

Training stability is not only a scientific question but a pressing practical concern for the safe and sustainable development of AI.
Unstable learning dynamics undermine reproducibility, exacerbate computational inequity through wasted resources, and can lead to unpredictable or hazardous system behaviors.
By providing tools to characterize and audit training stability, this work contributes to the development of more transparent, robust, and accountable AI systems.

\subsection{Future research directions}

\begin{enumerate}[leftmargin=2em]
    \item \textbf{Mechanistic interpretability of stability.}
    Why are some architectures or optimizers inherently more stable?
    Combining meta-state analysis with interpretability techniques could uncover causal pathways linking parameter geometry, optimization noise, and dynamical resilience.

    \item \textbf{Exploration of stability-aware interaction.}
    Building on our demonstration of conditional meta-state feedback, future work may investigate how meta-state-informed interaction can be integrated into training loops for scientific exploration.
    Our present results establish the \emph{monitoring-and-identifiability substrate} needed to make such questions well-posed, without asserting immediate deployment of stabilization strategies.

    \item \textbf{Expanding the empirical scope.}
    Applying StabilityBench to emerging paradigms such as diffusion models, vision transformers, multi-agent systems, meta-learning, and continual learning will test the generality of the principles identified here.

    \item \textbf{Stability-aware algorithm design.}
    Our taxonomy and metrics provide a foundation for designing algorithms with stability as a first-class objective, potentially leading to new optimizer families or regularization strategies that explicitly trade off capability for dynamical reliability.
\end{enumerate}

\subsection{Conclusion}

As deep learning systems grow in scale, autonomy, and societal impact, understanding and characterizing their learning dynamics will become as important as improving their final capabilities.
This work reframes training stability as a foundational scientific property and provides the conceptual, methodological, and empirical tools to study it systematically.
We hope this foundation catalyzes a broader shift toward building AI systems that are not only capable, but also robust, reliable, and scientifically interpretable.

\section{Methods}
\label{sec:methods}
StabilityBench implements perturbation auditing as a modular scientific instrument that decouples tasks, learners, perturbations, and measurements.

To ensure reproducibility and facilitate independent validation, we summarize here the core experimental protocols and implementation choices underlying StabilityBench. Detailed hyperparameters, architectural specifications, and extended analyses are provided in the Supplementary Information.

\subsection{Tasks and environments}
\label{subsec:tasks}


\textbf{Reinforcement Learning.} Experiments use standard continuous-control benchmarks
from the MuJoCo suite \cite{brockman2016openaigym}. Each environment defines a Markov decision process with continuous
actions; performance is measured as the average undiscounted return over 1,000-step
evaluation episodes.

\textbf{Large Language Models.} We train GPT‑2 variants (124M, 355M, 774M parameters) on a causal language‑modeling objective using tokenized, shuffled text batches of length 1,024. Validation perplexity on a held‑out set serves as the performance metric. All training signals (gradients, losses, optimizer states) are logged per optimization step for dynamical analysis.

\subsection{Learners and training protocols}
\label{subsec:learners}

\textbf{Reinforcement Learning.} Four actor‑critic algorithms are evaluated: PPO \cite{schulman2017ppo}, TRPO \cite{schulman2015trpo}, SAC \cite{haarnoja2018sac}, and TD3 \cite{fujimoto2018td3}. Hyperparameters follow established implementations (see \SuppTabRLHyperparams). Each run trains for 1--3 million environment steps, without early stopping.

\textbf{Large Language Models.} All models are trained with AdamW \cite{loshchilov2019adamw} (weight decay \(10^{-4}\)), linear warmup (1,000 steps), and cosine learning‑rate decay \cite{loshchilov2017sgdr}. Gradient clipping (global norm 1.0) is applied unless perturbed. Training lasts 50,000--150,000 steps depending on model size.

\subsection{Perturbation protocols}
\label{subsec:perturbations}

Perturbations are applied mid‑training (20--40\% of total steps) along the four stability dimensions:
\begin{itemize}
    \item \textbf{Optimization:} Learning‑rate spikes, optimizer‑moment noise, gradient perturbations.
    \item \textbf{Environmental/Data:} Action/observation noise (RL); token corruption/batch‑distribution shifts (LLMs).
    \item \textbf{Parametric:} Gaussian weight noise, partial layer resets.
    \item \textbf{Learning‑Signal:} Reward transformations (RL); gradient distortions (LLMs).
\end{itemize}

Exact perturbation magnitudes and injection schedules are provided in \SuppTabPerturbationSpecs.

\subsection{Stability metrics}
\label{subsec:metrics-methods}

We quantify stability using two complementary types of indicators:

\textbf{Run-level evaluation metrics} capture aggregate behavior across training runs:
\begin{itemize}[leftmargin=2em,noitemsep]
    \item \textbf{Collapse Time ($T_c$):} Step at which performance irreversibly deteriorates after perturbation.
    \item \textbf{Meta-State Deviation (MSD):} Maximum $\ell_2$ deviation of the learned latent state induced by perturbation over a fixed horizon.
    \item \textbf{Spike Intensity Profile (SIP):} Short-term instability spike intensity computed from performance/instability telemetry over a sliding window.
\end{itemize}
These metrics capture complementary facets of stability and are not reducible to a single ranking.

\textbf{Online meta-state variables} describe the temporal evolution of training dynamics:
\begin{itemize}[leftmargin=2em,noitemsep]
    \item \textbf{Gradient coherence (\(x_{grad}\)):} Alignment of gradient directions across mini-batches.
    \item \textbf{Instability index (\(x_{inst}\)):} Variance of performance over a sliding window.
    \item \textbf{Performance trend (\(x_{gen}\)):} Smoothed performance trajectory.
    \item \textbf{State persistence (\(x_{mem}\)):} Memory of recent training states.
\end{itemize}

Precise metric definitions are listed in \SuppTabMetricDefs.

\subsection{Meta-state monitor implementation}
\label{subsec:metastate-methods}

The Meta‑State Monitor learns a low‑dimensional representation \(h_t\) of the training process using a recurrent state‑space model. It ingests telemetry streams (gradients, losses, optimizer statistics) and outputs a latent trajectory that summarizes dynamical health. The model is trained on unperturbed runs to capture normal dynamics; deviations from this baseline serve as indicators of instability emergence.

\subsection{Compute resources and reproducibility}
\label{subsec:compute}

RL experiments used 1--4 NVIDIA V100/A100 GPUs per run; LLM experiments used up to 8 GPUs. Complete code, configuration files, and logged telemetry are available in the project repository to ensure exact reproducibility.

\bibliographystyle{plainnat}
\bibliography{refs}

\end{document}


\maketitle



\newcommand{\SuppFigInstabilityPerturb}{Supplementary Fig.~S1}
\newcommand{\SuppFigArchitecture}{Supplementary Fig.~S2}
\newcommand{\SuppFigAdditionalLLM}{Supplementary Fig.~S3}


\newcommand{\SuppTabRLHyperparams}{Supplementary Table~S2}
\newcommand{\SuppTabPerturbationSpecs}{Supplementary Table~S4}
\newcommand{\SuppTabMetricDefs}{Supplementary Table~S3}


\newcommand{\SuppSecArchitecture}{Supplementary Section~S3}



\newcommand{\MainFigRLSensitivity}{Fig.~2}

\setcounter{section}{0}
\renewcommand{\thesection}{S\arabic{section}}

\setcounter{figure}{0}
\setcounter{table}{0}
\renewcommand{\thefigure}{S\arabic{figure}}
\renewcommand{\thetable}{S\arabic{table}}

\section{Meta-state representation and structured aggregation}
\label{sec:meta-state}

To characterize training dynamics beyond surface performance, we introduce a structured meta-state representation that aggregates multiple channels of training telemetry. This representation provides a unified, low-dimensional summary of training dynamics, enabling systematic analysis and monitoring across learning paradigms.

\subsection{Meta-state representation: construction and motivation}

\begin{figure}[t]
    \centering
    \includegraphics[width=0.98\linewidth]{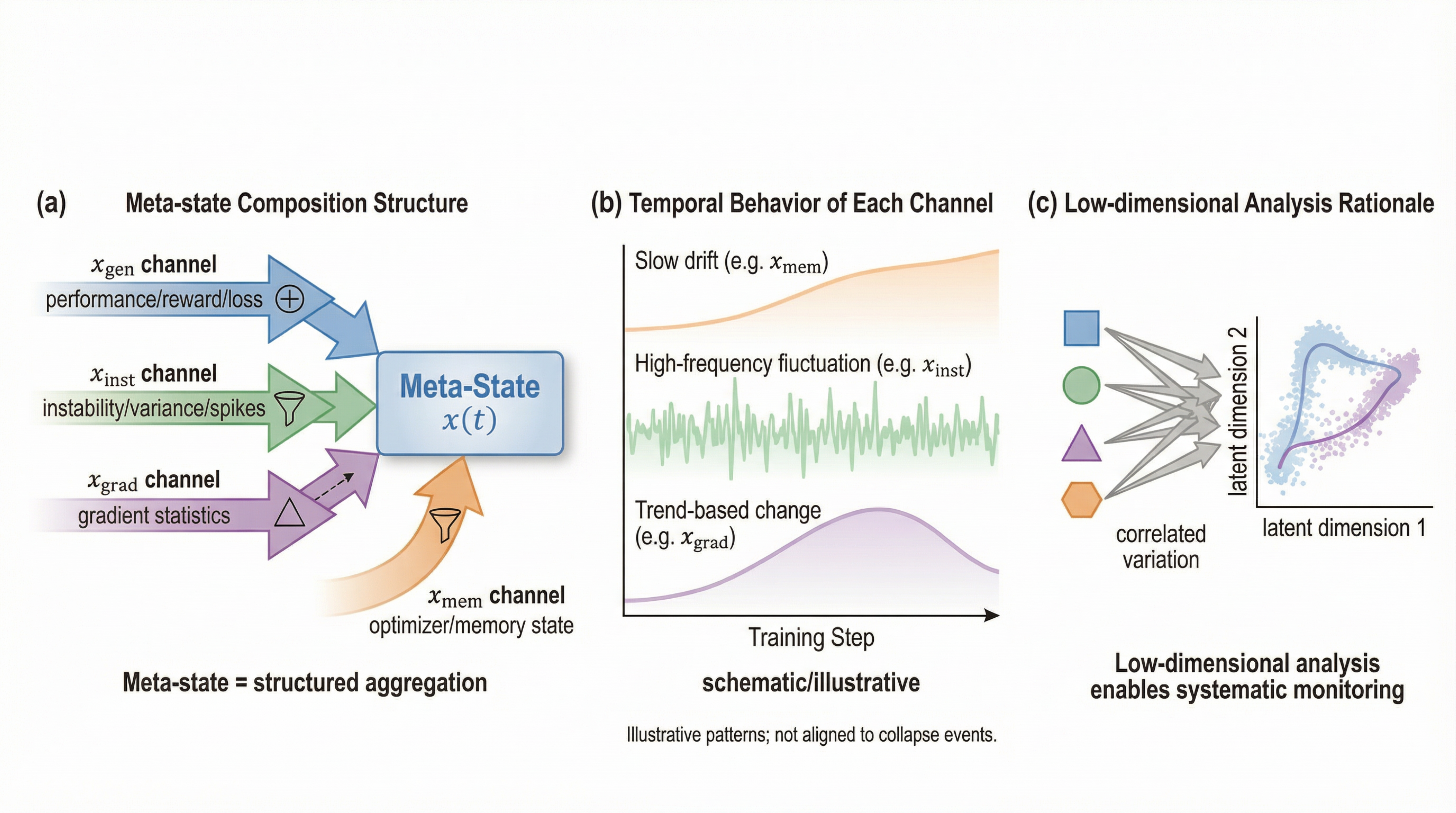}
    \caption{Schematic illustration of the meta-state representation and its analytical motivation.
(a) Meta-state composition structure, showing how multiple structured channels derived from training telemetry—performance-related signals ($x_{\mathrm{gen}}$), instability indicators ($x_{\mathrm{inst}}$), gradient statistics ($x_{\mathrm{grad}}$), and optimizer or memory state ($x_{\mathrm{mem}}$)—are aggregated into a unified meta-state representation $x(t)$.
(b) Illustrative temporal behavior of individual channels, highlighting their distinct characteristic dynamics across training, including slow drift, high-frequency fluctuations, and trend-based changes. These patterns are schematic and are not aligned to collapse events.
(c) Conceptual rationale for low-dimensional analysis, illustrating how correlated variation across channels motivates the use of compact latent summaries for systematic monitoring of training dynamics.
}
    \label{fig:phase}
\end{figure}

Figure~\ref{fig:phase} provides a schematic overview of the meta-state representation used throughout this work. The meta-state aggregates multiple structured channels derived from training telemetry, including performance-related signals, instability indicators, gradient statistics, and optimizer or memory state. This construction is designed to capture complementary aspects of training dynamics across multiple time scales, while remaining interpretable and grounded in observable quantities.

The figure further illustrates the characteristic temporal behavior of each channel in a schematic manner, highlighting their distinct roles in capturing slow drift, high-frequency fluctuations, and trend-based changes during training. Finally, the schematic demonstrates why correlated variation across these channels motivates the use of low-dimensional latent summaries for systematic monitoring and analysis. All visual patterns are illustrative and are not aligned to specific collapse events.

\subsection{Latent meta-state monitoring and representation}

To characterize instability before it becomes visible in performance metrics, we incorporate a latent meta-state representation of the training process. High-dimensional telemetry streams are mapped into a low-dimensional latent state \(h_t\) using a learned state-space model, which summarizes the dynamical health of the system. The relationship between latent-state deviations and performance outcomes is analyzed in the main text (Section 2.3).

\subsection{Stability metrics}

Stability is quantified using a suite of complementary dynamical metrics, computed alongside standard performance curves. These include collapse time, short-term instability indices, the probability of divergence across random seeds, and deviations in the latent meta-state following perturbation. Together, these measures allow transient noise to be distinguished from irreversible dynamical failure and enable systematic comparison of stability across algorithms and domains.

Implementation details, perturbation schedules, metric definitions, and the StabilityBench system architecture are provided in Methods and Supplementary Section~\ref{si:architecture}.

\section{Robustness and generalization analysis of meta-state deviation}
\label{si:robustness}

\begin{figure}[htbp]
    \centering
    \includegraphics[width=0.48\linewidth]{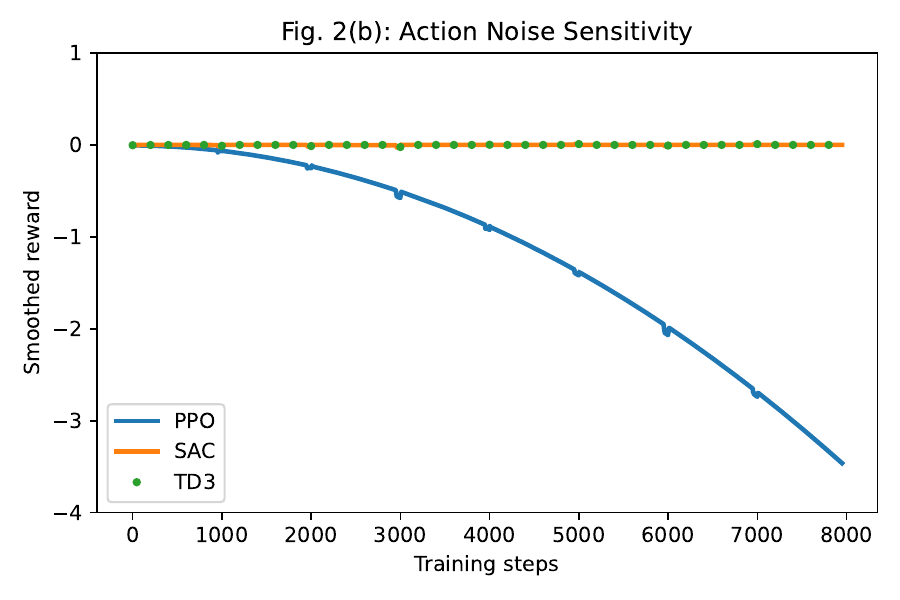}
    \hfill
    \includegraphics[width=0.48\linewidth]{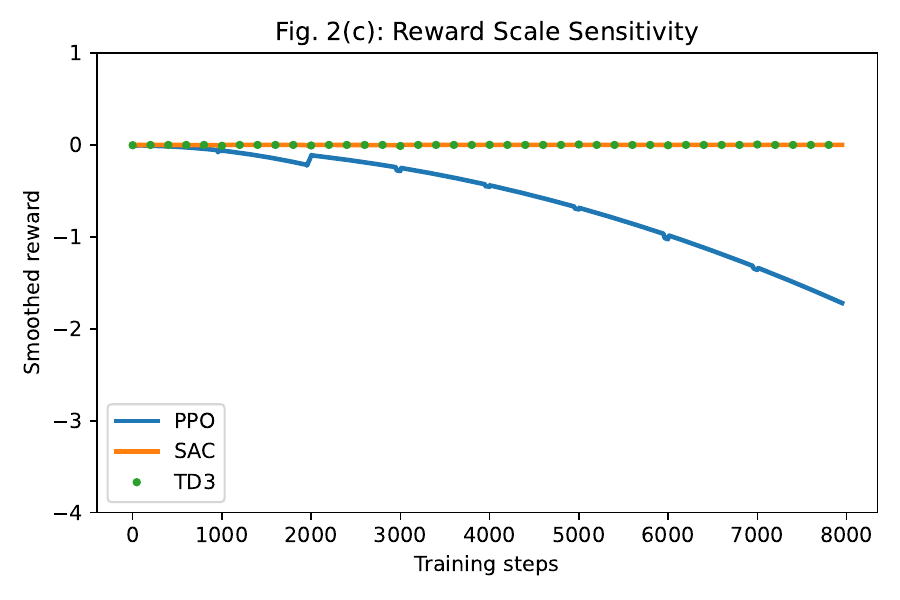}
    \caption{
    \textbf{Additional perturbation analyses supporting algorithm-specific instability.}
    \textbf{(a)} Sensitivity to continuous action noise.
    PPO exhibits sustained performance deterioration under persistent action perturbations,
    whereas SAC and TD3 maintain stable learning dynamics.
    \textbf{(b)} Sensitivity to reward-scale perturbation.
    PPO shows pronounced degradation under reward scaling,
    while off-policy methods remain robust.
    Together, these analyses indicate that the catastrophic instability observed in \MainFigRLSensitivity{} extends beyond learning-rate perturbations, reflecting a broader algorithm-dependent sensitivity across multiple perturbation dimensions.
    }
    \label{fig:supp_instability_perturb} 
\end{figure}

To assess the generality of the algorithm-specific instability observed in \MainFigRLSensitivity{},
we further examine training dynamics under additional perturbation modalities (Supplementary Figure ~\ref{fig:supp_instability_perturb}).
Specifically, we consider continuous action noise and reward-scale perturbations,
which probe complementary sources of stochasticity and signal distortion
beyond learning-rate manipulation.

This section examines the distributional behavior of meta-state deviation across random seeds, model scales, and training outcomes. The analysis focuses on distinguishing collapse runs from stable runs, demonstrating that meta-state deviation exhibits consistent and statistically significant separation across perturbation types and experimental conditions. All statistics are computed over pre-collapse windows, consistent with the main analysis.

The analysis further confirms that meta-state deviation serves as a reliable and domain-agnostic indicator of training fragility, supporting its use as a core stability metric in the StabilityBench framework.

\begin{figure}[t]
    \centering
    \includegraphics[width=0.98\linewidth]{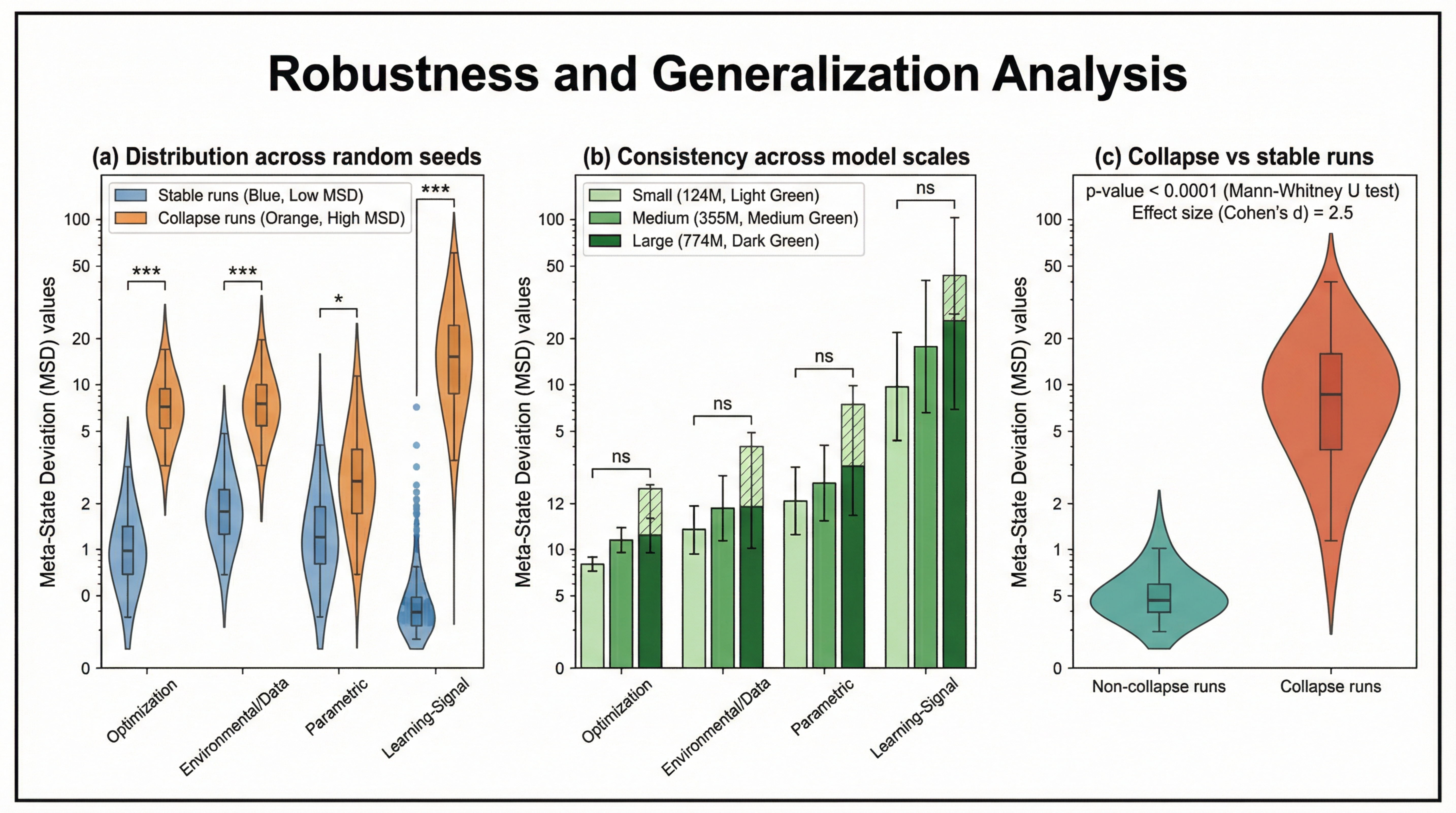}
    \caption{Robustness and generalization analysis of meta-state deviation across experimental conditions. (a) Distribution of meta-state deviation for collapse vs. non-collapse runs across multiple random seeds. (b) Meta-state deviation as a function of model scale (parameters), showing consistent separation. (c) Comparison of deviation magnitudes across different perturbation types, highlighting the metric’s sensitivity to instability dimensions. All statistics are computed over pre-collapse windows; error bars indicate standard error across seeds.}
    \label{fig:robustness}
\end{figure}

\section{Implementation details of the StabilityBench auditing system}
\label{si:architecture}

This section documents the architectural and implementation choices underlying StabilityBench. The system is designed to enable controlled perturbation auditing across learning paradigms.

\subsection{Architectural overview}
\label{subsec:architecture}

The StabilityBench framework is built around a modular decomposition of training dynamics into four layers: task, learner, perturbation, and measurement. This decomposition allows the same stability auditing protocol to be instantiated across reinforcement learning and large language model training. An \texttt{AuditRunner} operationalizes this abstraction by orchestrating execution across layers: it runs training under controlled perturbation schedules, logs comprehensive telemetry, and ensures full reproducibility through strict seeding and checkpointing.

\subsection{Latent-state monitoring and declarative auditing}
\label{subsec:metastate_and_api}

StabilityBench incorporates a \textbf{Meta-State Monitor} that learns a low-dimensional representation of the training process. It maps high-dimensional telemetry streams (gradients, losses, optimizer statistics) into latent coordinates \(h_t\) via a state-space model:
\begin{equation}
h_{t+1} = f_\phi(h_t, y_t),
\end{equation}
where \(y_t\) is the observed telemetry.

A key design feature is \emph{declarative auditing}. Users specify \emph{what} to perturb (e.g., ``optimization dimension''), \emph{how} strongly, and \emph{when}; StabilityBench handles the domain-specific implementation. For example, a ``learning-rate spike'' perturbation is automatically applied to the optimizer of a PPO agent or an LLM training run. This abstraction enables controlled, cross-paradigm stress-testing that would be cumbersome to implement manually.

\subsection{Quantifying stability}
\label{subsec:metrics}

StabilityBench operationalizes training stability through a set of complementary dynamical measurements. The measurement layer integrates multiple signals that capture both surface-level performance degradation and latent dynamical deviation.

Concretely, stability is assessed along four complementary axes, drawing on established benchmarking principles for systematic evaluation of model robustness under perturbation.

Formal definitions, thresholds, and computational details of each metric are provided in Methods (Section~5.4) and Supplementary Table~\ref{tab:metric_defs}.

\subsection{Extensibility and reproducibility}
\label{subsec:extensibility}

The framework is built for extension: new tasks, learners, perturbation types, and metrics can be registered through a uniform API. Every audit is fully reproducible—perturbation schedules, random seeds, telemetry logs, and latent trajectories are serialized, enabling exact replay and secondary analysis.

\section{Extended analyses and ablation studies}
\label{si:expdetails}

This section provides additional methodological details and ablation studies that support the main experimental protocols.

\subsection{Extended taxonomy details}
\label{si:taxonomy_extended}

\subsubsection{Four-dimensional stability taxonomy: extended discussion}

Our four-dimensional taxonomy categorizes training instabilities based on which component of the learning process is perturbed. Below we provide extended descriptions and concrete examples across reinforcement learning (RL) and large language model (LLM) training.

\paragraph{Optimization stability.}
This dimension captures sensitivity to perturbations in the update mechanism itself.

\begin{itemize}
    \item \textbf{RL examples:} Learning-rate spikes; momentum buffer corruption.
    \item \textbf{LLM examples:} Sudden changes in Adam's second-moment estimates; gradient clipping threshold violations.
\end{itemize}

\paragraph{Environmental / data stability.}
This dimension measures robustness to changes in the data-generating process.

\begin{itemize}
    \item \textbf{RL examples:} Observation noise; non-stationary dynamics.
    \item \textbf{LLM examples:} Token corruption; batch distribution drift.
\end{itemize}

\paragraph{Parametric stability.}
This dimension addresses the geometry of the parameter landscape.

\begin{itemize}
    \item \textbf{RL examples:} Policy weight noise; critic resets.
    \item \textbf{LLM examples:} Weight noise during fine-tuning; layer re-initialization.
\end{itemize}

\paragraph{Learning-signal stability.}
This dimension concerns the fidelity of the signal that guides learning.

\begin{itemize}
    \item \textbf{RL examples:} Reward shaping errors; adversarial reward perturbations.
    \item \textbf{LLM examples:} Gradient sign flips; objective perturbations.
\end{itemize}

\subsubsection{Cross-domain instantiations}

Table~\ref{tab:taxonomy_instances} summarizes how each stability dimension manifests in RL and LLM training.

\subsection{Ablation studies}
\label{si:ablations}

\subsubsection{Hyperparameter settings}

\paragraph{Reinforcement learning.}
Table~\ref{tab:rl_hyperparams} lists the hyperparameters for all RL algorithms used in our experiments.

\paragraph{Large language models.}
Table~\ref{tab:llm_hyperparams} details the training configurations for GPT-2 variants.

\subsubsection{Perturbation magnitudes and schedules}

Exact perturbation magnitudes and injection schedules for all stability dimensions are provided in Table~\ref{tab:perturbation_specs}.

\subsubsection{Ablation results}

\paragraph{Effect of perturbation timing.}
We varied the injection time \(t_s\) from 10\% to 90\% of training. Early perturbations (10--30\%) were often recoverable, while late perturbations (70--90\%) frequently caused irreversible collapse. Mid-training perturbations (30--50\%) provided the clearest discrimination between stable and unstable algorithms.

\paragraph{Metric sensitivity analysis.}
Collapse time and meta-state deviation exhibited strong alignment across runs. Recovery rate provided complementary signal for partial failures.

\paragraph{Algorithm-specific sensitivities.}
Within RL, on-policy methods (PPO, TRPO) were most sensitive to environmental perturbations, while off-policy methods (SAC, TD3) were more robust to data shifts but more sensitive to reward corruption. In LLMs, larger models exhibited disproportionate fragility to optimization perturbations.

\paragraph{Cross-paradigm transfer.}
Algorithms that maintained entropy in RL also maintained gradient coherence in LLM training.

\section{Supplementary tables and auditing specifications}
\label{si:tables}

This appendix provides supplementary tables documenting experimental configurations, perturbation protocols, metric definitions, and compute resources.

\begin{table*}[htbp]
\centering
\caption{\textbf{Cross–paradigm manifestations of training instability.}}
\small
\setlength{\tabcolsep}{6pt}
\renewcommand{\arraystretch}{1.2}
\begin{tabularx}{\textwidth}{lXX}
\toprule
\textbf{Dimension} 
& \multicolumn{1}{c}{\textbf{RL Instantiations}} 
& \multicolumn{1}{c}{\textbf{LLM Instantiations}} \\
\midrule
Optimization 
& Learning-rate spikes; momentum corruption; gradient clipping failures 
& Adam moment drift; learning-rate schedule errors; clipping violations \\

Environmental / Data 
& Observation noise; action perturbations; non-stationary dynamics 
& Token corruption; batch distribution drift; streaming data shifts \\

Parametric 
& Policy weight noise; critic resets; quantization effects 
& Transformer weight perturbations; layer re-initialization; mixed precision \\

Learning-Signal 
& Reward shaping errors; adversarial rewards; delayed signals 
& Gradient sign flips; objective perturbations; gradient masking \\
\bottomrule
\end{tabularx}
\label{tab:taxonomy_instances}
\end{table*}


\begin{table*}[htbp]
\centering
\caption{Reinforcement learning hyperparameters.}
\begin{tabularx}{\textwidth}{lXXXX}
\toprule
\textbf{Hyperparameter} & \textbf{PPO} & \textbf{TRPO} & \textbf{SAC} & \textbf{TD3} \\
\midrule
Learning rate & $3\times10^{-4}$ & $1\times10^{-3}$ & $3\times10^{-4}$ & $3\times10^{-4}$ \\
Batch size & 64 & 64 & 256 & 256 \\
Replay buffer size & -- & -- & $10^6$ & $10^6$ \\
Discount factor ($\gamma$) & 0.99 & 0.99 & 0.99 & 0.99 \\
Target update rate ($\tau$) & -- & -- & $5\times10^{-3}$ & $5\times10^{-3}$ \\
Entropy coefficient & 0.01 & -- & 0.2 & -- \\
GAE parameter ($\lambda$) & 0.95 & 0.95 & -- & -- \\
Clipping parameter & 0.2 & -- & -- & -- \\
Trust region size & -- & 0.01 & -- & -- \\
\bottomrule
\end{tabularx}
\label{tab:rl_hyperparams}
\end{table*}

\begin{table*}[htbp]
\centering
\caption{\textbf{Hyperparameter settings for large language model training.}}
\small
\setlength{\tabcolsep}{5pt}
\renewcommand{\arraystretch}{1.15}
\begin{tabularx}{\textwidth}{l c c c X}
\toprule
\textbf{Model size} 
& \textbf{Learning rate} 
& \textbf{Batch size} 
& \textbf{Sequence length} 
& \textbf{Optimizer details} \\
\midrule
124M 
& $5\times10^{-4}$ 
& 32 
& 1024 
& AdamW ($\beta_1=0.9$, $\beta_2=0.95$), clip $=1.0$ \\
355M 
& $5\times10^{-4}$ 
& 32 
& 1024 
& AdamW ($\beta_1=0.9$, $\beta_2=0.95$), clip $=1.0$ \\
774M 
& $5\times10^{-4}$ 
& 32 
& 1024 
& AdamW ($\beta_1=0.9$, $\beta_2=0.95$), clip $=1.0$ \\
\bottomrule
\end{tabularx}
\label{tab:llm_hyperparams}
\end{table*}

\begin{table*}[htbp]
\centering
\caption{\textbf{Standardized perturbation protocols for stability auditing.}}
\small
\setlength{\tabcolsep}{6pt}
\renewcommand{\arraystretch}{1.2}
\begin{tabularx}{\textwidth}{l l l p{0.35\textwidth}}
\toprule
\textbf{Dimension} & \textbf{Perturbation} & \textbf{Magnitude} & \textbf{Protocol} \\
\midrule
\multirow{3}{*}{Optimization}
& LR spike & $10\times$ & 10 steps at 30\% training progress \\
& Momentum noise & $\mathcal{N}(0,0.1)$ & Gaussian noise on optimizer momentum \\
& Gradient scaling & $\pm20\%$ & Random gradient rescaling \\
\midrule
\multirow{3}{*}{Environmental / Data}
& Observation noise (RL) & $\mathcal{N}(0,0.1)$ & Additive state noise \\
& Action noise (RL) & $\mathcal{N}(0,0.05)$ & Additive action noise \\
& Token corruption (LLM) & 5\% & Random token masking \\
\midrule
\multirow{2}{*}{Parametric}
& Weight noise & $\mathcal{N}(0,0.01)$ & Additive parameter noise \\
& Layer reset & 1 layer & Random layer re-initialization \\
\midrule
\multirow{3}{*}{Learning-Signal}
& Reward noise (RL) & $\mathcal{N}(0,0.5)$ & Additive reward perturbation \\
& Gradient sign flip (LLM) & 10\% & Random sign inversion \\
& Objective perturbation (LLM) & $\alpha=0.2$ & Label smoothing \\
\bottomrule
\end{tabularx}
\label{tab:perturbation_specs}
\end{table*}

\begin{table*}[htbp]
\centering
\caption{\textbf{Stability metrics used in StabilityBench.}}
\small
\setlength{\tabcolsep}{6pt}
\renewcommand{\arraystretch}{1.15}
\begin{tabularx}{0.9\textwidth}{l l X}
\toprule
\textbf{Metric} & \textbf{Notation} & \textbf{Concept and formal definition} \\
\midrule
Collapse Time 
& $T_c$ 
& \textit{Time to irreversible performance collapse.}
First step $t$ such that $\mathcal{J}(t) < \mathcal{J}_{\text{pre}} - 2\sigma_{\text{pre}}$
and remains below this threshold for $\Delta=100$ steps. \\

Instability Index 
& $I_{\text{inst}}$ 
& \textit{Short-term fluctuation intensity.}
$I_{\text{inst}} = \frac{1}{W}\sum_{i=t-W+1}^{t} (\mathcal{J}(i) - \bar{\mathcal{J}}_W)^2$,
with sliding window $W=50$. \\

Divergence Probability 
& $P_{\text{div}}$ 
& \textit{Probability of early catastrophic failure.}
Fraction of random seeds for which $T_c < T_{\max}/2$. \\

Recovery Rate 
& $R_{\text{rec}}$ 
& \textit{Ability to recover after perturbation.}
$R_{\text{rec}} = \frac{\mathcal{J}(t_{\text{end}}) - \mathcal{J}(t_{\text{min}})}
{\mathcal{J}_{\text{pre}} - \mathcal{J}(t_{\text{min}})}$
for runs that do not fully collapse. \\

Meta-State Deviation
& \textbf{MSD} ($D_{\text{meta}}$)
& \textit{Maximum latent deviation induced by perturbation.}
$D_{\text{meta}} = \max_{t \in [t_s, t_s+T]} \|h_t - h_{t_s}\|_2$, with $T=500$. \\
\bottomrule
\end{tabularx}
\label{tab:metric_defs}
\end{table*}

\begin{table*}[htbp]
\centering
\caption{\textbf{Compute resources used in StabilityBench experiments.}}
\small
\setlength{\tabcolsep}{8pt}
\renewcommand{\arraystretch}{1.15}
\begin{tabularx}{0.9\textwidth}{
>{\raggedright\arraybackslash}X
>{\centering\arraybackslash}p{0.2\textwidth}
>{\centering\arraybackslash}p{0.15\textwidth}
}
\toprule
\textbf{Experiment component} & \textbf{GPU hours} & \textbf{Runs} \\
\midrule
RL baseline training & 2,400 & 800 \\
RL perturbation audits & 7,200 & 2,400 \\
LLM baseline training & 3,600 & 300 \\
LLM perturbation audits & 10,800 & 900 \\
Meta-state monitor training & 1,200 & N/A \\
\midrule
\textbf{Total} & \textbf{25,200} & \textbf{4,400} \\
\bottomrule
\end{tabularx}
\label{tab:compute}
\end{table*}

\FloatBarrier